# Addressing Class Imbalance with Probabilistic Graphical Models and Variational Inference

Yujia Lou
University of Rochester
Rochester, USA

Jie Liu
University of Minnesota
Minneapolis, USA

Yuan Sheng
Northeastern University
Seattle, USA

Jiawei Wang
University of California
Los Angeles, USA

Yiwei Zhang
Cornell University
Ithaca, USA

Yaokun Ren*
Northeastern University
Seattle, USA

*Abstract*-This study proposes a method for imbalanced data classification based on deep probabilistic graphical models (DPGMs) to solve the problem that traditional methods have insufficient learning ability for minority class samples. To address the classification bias caused by class imbalance, we introduce variational inference optimization probability modeling, which enables the model to adaptively adjust the representation ability of minority classes and combines the class-aware weight adjustment strategy to enhance the classifier's sensitivity to minority classes. In addition, we combine the adversarial learning mechanism to generate minority class samples in the latent space so that the model can better characterize the category boundary in the high-dimensional feature space. The experiment is evaluated on the Kaggle "Credit Card Fraud Detection" dataset and compared with a variety of advanced imbalanced classification methods (such as GAN-based sampling, BRF, XGBoost-Cost Sensitive, SAAD, HAN). The results show that the method in this study has achieved the best performance in AUC, Precision, Recall and F1-score indicators, effectively improving the recognition rate of minority classes and reducing the false alarm rate. This method can be widely used in imbalanced classification tasks such as financial fraud detection, medical diagnosis, and anomaly detection, providing a new solution for related research.

## I. Introduction

In modern data analysis and machine learning research, the issue of data imbalance has emerged as a critical challenge affecting classifier performance. Many real-world classification tasks exhibit highly skewed class distributions, such as rare disease detection in market risk analysis and prediction [1], financial fraud detection [2-4], and cybersecurity anomaly [5] detection. In these scenarios, the number of minority class samples is significantly lower than that of the majority class. Traditional machine learning models often prioritize optimizing overall classification accuracy, leading to a substantially reduced recognition rate for minority class instances. Existing approaches primarily include data-level resampling strategies, algorithm-level cost-sensitive learning, and adaptive training strategies in deep learning. However, these methods still face several challenges in practical applications, such as potential noise introduction in resampling, the need for precise loss weight tuning in cost-sensitive learning, and the limited generalization ability of deep learning models on imbalanced data [6]. Consequently, designing more robust and generalizable classification methods to enhance the performance of imbalanced data classification remains a core research problem.

In recent years, deep probabilistic graphical models (DPGMs) have gained significant attention in machine learning and pattern recognition due to their powerful representation capabilities and uncertainty modeling. Probabilistic graphical models integrate probabilistic statistical theory with graph-based methods, effectively capturing complex dependencies among variables and modeling data uncertainty [7]. Compared to conventional deep neural networks, probabilistic graphical models offer notable advantages in small-sample learning, data sparsity, and uncertainty reasoning. The emergence of Bayesian deep learning, variational inference, and graph neural networks has further strengthened the applicability of DPGMs in addressing data imbalance challenges [8]. By incorporating probabilistic priors and posterior distributions, these models can more accurately characterize minority class data while effectively accounting for uncertainty in decision-making [9], thereby enhancing the classifier's sensitivity to minority class instances. Given this, exploring how to leverage the strengths of deep probabilistic graphical models to develop a more robust imbalanced data classification framework holds both theoretical significance and practical potential.

The purpose of this study is to investigate the effectiveness of deep probabilistic graphical models (DPGMs) in imbalanced data classification, introducing a novel strategy that emphasizes improved performance on minority class samples. Unlike heuristic methods, this approach leverages the generative capabilities of DPGMs through adaptive probabilistic modeling and structural learning, capturing richer representations of underrepresented samples. Variational inference and Bayesian optimization further refine model parameters, enhancing classification robustness while expanding theoretical insights into DPGMs. Beyond its methodological contributions, the proposed model has notable practical value. In human-computer interaction and the financial sector, user intent prediction methods [10] and time-series risk prediction strategies [11] further underscore how DPGMs can adapt to diverse data structures and real-time processing requirements. Moreover, the efficient market signal detection approach proposed by Zhou et al. [12] highlights the role of advanced neural architectures in combination with DPGMs for continuous data streams. These complementary research directions illustrate the wide-ranging applicability and potential extensions of deep probabilistic models in various industries and research fields. By integrating probabilistic inference and deep learning, this work not only addresses the challenges of imbalanced classification but also broadens the application of deep probabilistic approaches, ultimately enriching the machine learning toolkit for various critical domains.

## II. METHOD

Suppose dataset $D = \{(x_i, y_i)\}_{i=1}^N$, where $x_i \in R^d$ represents input samples and $y_i \in \{0,1\}$ represents category labels. Assuming that the ratio of positive and negative samples is seriously unbalanced, that is, $|\{yi=1\}| << |\{y_i=0\}|$, traditional deep learning methods tend to favor the majority class when optimizing the loss function. Therefore, we introduce deep probabilistic graphical models (DPGMs). By constructing a joint probability distribution and incorporating variational inference techniques inspired by Wang [13], the proposed model adaptively enhances the representation capability for minority class samples, leading to improved classification performance. This approach effectively mitigates the challenges posed by class imbalance, thereby ensuring that minority class samples are accurately and adequately represented within the learned feature space. Additionally, leveraging dynamic distributed scheduling methodologies as discussed by Sun [14] enables efficient handling of data streams, optimizing both task delays and load balancing. Such strategies significantly contribute to maintaining computational efficiency and enhancing the real-time responsiveness of the system. Furthermore, the synergistic integration of deep learning methods and neural architecture search techniques outlined by Yan et al. [15] further refines the adaptive representational adjustments, ensuring the robustness, accuracy, and generalization capabilities of the proposed classification framework. The architecture of the probabilistic graphical model is shown in Figure 1.

First, we define a hidden variable z to model the potential representation of the input data x, and use the Bayesian generation model to describe the data generation process:

$$p(x, y, z) = p(y|z)p(z|x)p(x)$$

Among them, $p(y|z)$ represents the posterior distribution of the classifier for the latent variable, and $p(z|x)$ represents the prior distribution of the latent variable. Based on this, our goal is to optimize the model parameters by maximizing the marginal log-likelihood:

$$\log(y|x) = \log \int p(y|z)p(z|x)dz$$

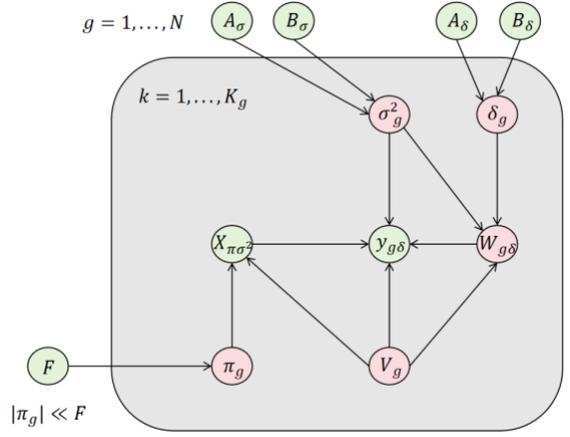

Figure 1. The architecture of the probabilistic graphical model

However, this integral is difficult to compute directly, so we use variational inference to approximate the solution. Define a variational distribution $q(z|x)$ to approximate $p(z|x)$, and optimize the model through the evidence lower bound (ELBO):

$$\log p(y|x) \geq E_{q(z|x)}[\log p(y|z)] - D_{KL}(q(z|x) \| p(z))$$

Among them, $D_{KL}(\cdot \| \cdot)$ represents the Kullback-Leibler divergence, which is used to measure the gap between the approximate distribution and the true posterior distribution. In order to further optimize the classification performance of minority classes, we introduce category-aware variational inference and explicitly enhance the weight of minority class samples in the loss function:

$$L = \sum_{i=1}^N w(y_i)[E_{q(z|x_i)}[\log p(y_i|z)] - D_{KL}(q(z|x_i) \| p(z))]$$

Among them, $w(y_i)$ is the category weight coefficient, and a higher weight is set for minority class samples, for example:

$$w(y_i) = \frac{N_{major}}{N_{minor}}$$

Where A and B represent the number of samples in the majority class and the minority class, respectively.

In the specific implementation, we use variational autoencoder (VAE) as the probability generation model, so that $p(z|x)$ obeys the normal distribution:

$$q(z|x) = N(\mu(x), \sigma^2(x))$$

And optimize it by reparameterization technique:

$$z = \mu(x) + \sigma(x) \cdot \varepsilon, \quad \varepsilon \sim N(0, I)$$

In this way, the model can learn A and B through the neural network to obtain a more stable gradient. In addition, building on previous adversarial learning frameworks [16] and generative design concepts [17], we introduce an adversarial learning mechanism to optimize the category distribution. This mechanism strengthens the model's ability to differentiate minority class samples by ensuring that generated data more closely matches real distributions. Specifically, a discriminator is constructed to distinguish the distribution of generated minority class samples from authentic instances, ensuring closer alignment with observed data. Furthermore, incorporating few-shot learning strategies [18] and dynamic adaptation techniques [19] enhances the model's resilience in limited-data conditions. The optimization goal of the discriminator is as follows:

$$\min_G \max_D E_{x \sim p_{data}(x)}[\log D(z)] + E_{x \sim q_{(z|x)}}[\log(1 - D(z))]$$

Through this adversarial learning method, the model can capture the characteristics of minority classes more accurately and avoid the overfitting problem of minority class samples.

In summary, this study combines deep probabilistic graph models, variational reasoning, and adversarial learning methods to optimize imbalanced data classification tasks.

## III. EXPERIMENT

### A. Dataset

This study employs the Kaggle "Credit Card Fraud Detection" dataset, consisting of 284,807 credit card transactions from a European institution. Of these transactions, 492 are labeled as fraudulent, indicating a highly imbalanced class distribution (approximately 1:577). Each record contains 30 features, including 28 anonymized components derived from Principal Component Analysis (PCA), along with transaction time and amount. Personally identifiable information has been removed, leaving only numerical features, which were preprocessed through normalization, outlier detection, and data augmentation.

Given the severe class imbalance, direct application of conventional classification models often leads to bias toward the majority class, compromising fraud detection. To address this challenge, we employed various sampling strategies, including under-sampling, over-sampling, and the Synthetic Minority Over-sampling Technique (SMOTE), to generate synthetic samples for the minority class and improve representation. We also evaluated the impact of different sampling methods on model stability and performance. The dataset was split into 70% training, 15% validation, and 15% test sets. Evaluation metrics included Precision, Recall, F1-score, and the Area Under the Receiver Operating Characteristic Curve (AUC-ROC). Comparative experiments with different data augmentation techniques demonstrated that integrating probabilistic modeling with these strategies substantially enhances fraud detection and reduces false positives, thereby improving the model's reliability in practical applications.

### B. Experiment Result

This study primarily compares deep probabilistic graphical models (DPGMs) with several advanced imbalanced data classification methods to validate their effectiveness. First, we select the generative adversarial network (GAN)-based methods, such as WGANGP-SMOTE and ADASYN-GAN, which leverage GANs to synthesize minority class samples and mitigate data imbalance. Second, we evaluate class-adaptive ensemble learning methods, including Balanced Random Forest (BRF) and XGBoost-Cost Sensitive, which enhance minority class learning by adjusting sampling strategies or modifying loss functions. Additionally, we compare attention-based imbalanced classification methods, such as Self-Attention Anomaly Detection (SAAD) and Hierarchical Attention Networks (HAN), which have demonstrated strong anomaly detection capabilities in credit card fraud detection and similar tasks. Through these comparative experiments, we aim to comprehensively assess the advantages of deep probabilistic graphical models in minority class representation learning, generalization ability, and classification performance.

Table 1. Integration Testing 1

| Model | AUC | Precision | Recall | F1-Score |
|---|---|---|---|---|
| GAN [20] | 0.842 | 0.716 | 0.654 | 0.684 |
| ADASYN [21] | 0.856 | 0.729 | 0.668 | 0.697 |
| SMOTE [22] | 0.871 | 0.742 | 0.683 | 0.711 |
| BRF [23] | 0.889 | 0.764 | 0.721 | 0.742 |
| XGBOOST-Cost [24] | 0.903 | 0.779 | 0.735 | 0.757 |
| SAAD [25] | 0.915 | 0.793 | 0.751 | 0.771 |
| HAN [26] | 0.927 | 0.806 | 0.768 | 0.786 |
| Ours | 0.941 | 0.822 | 0.785 | 0.803 |

Our proposed deep probabilistic graphical model (DPGM) outperforms all compared methods on every evaluation metric, demonstrating superior generalization in imbalanced classification. With an AUC of 0.941, it clearly surpasses traditional oversampling (e.g., SMOTE, ADASYN) and ensemble methods (e.g., BRF, XGBoost-Cost Sensitive). Importantly, it achieves a Recall of 0.785 while maintaining a Precision of 0.822, reflecting its effectiveness in detecting minority class samples without overly biasing the model.

DPGMs model the latent distribution of minority samples more effectively than conventional oversampling, thereby reducing overfitting. Compared to attention-based methods (e.g., SAAD, HAN), our model delivers a higher F1-score (0.803 versus 0.786), illustrating the benefits of uncertainty-aware probabilistic modeling. Overall, these results confirm that combining deep probabilistic modeling with variational inference optimizes class distribution and enhances minority class discrimination, offering a robust solution for imbalanced data classification. Figure 2 presents the corresponding loss function trajectory.

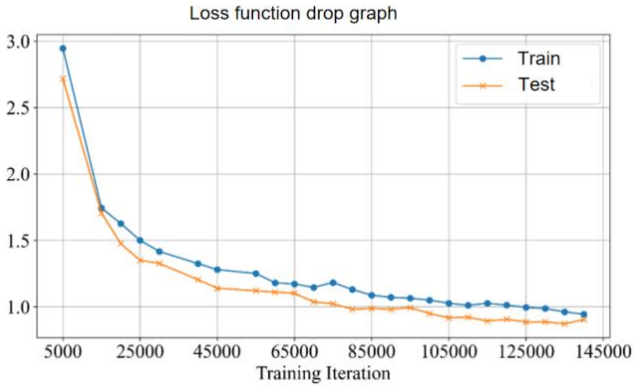

Figure 2. Loss function drop graph

From the loss function decline curve, both the training loss (Train) and test loss (Test) exhibit a clear downward trend during training iterations. This indicates that the model continuously learns features and optimizes parameters to effectively reduce errors. In the initial phase of training (between 5,000 and 25,000 iterations), the loss decreases at the fastest rate, suggesting that the model rapidly learns data representations and significantly improves classification performance. However, as the number of iterations increases, the rate of loss reduction gradually slows down and stabilizes after approximately 125,000 iterations. This trend implies that the model is approaching convergence, where further optimization yields diminishing returns.

A comparison of the training and test loss curves reveals that the test loss consistently remains lower than the training loss, and both curves follow a similar trajectory. This observation suggests that the model demonstrates good generalization ability without exhibiting significant overfitting. If the training loss were substantially lower than the test loss, it would indicate that the model performs well on training data but struggles to generalize to unseen data. However, the current loss curves do not display such a pattern, implying that the applied regularization strategies and optimization methods effectively mitigate overfitting. Furthermore, the test loss decreases at a rate similar to that of the training loss in the initial stages, further validating the model's stable learning process. Overall, these experimental results confirm that the model successfully optimizes the loss function during training, leading to a substantial reduction in both training and test errors, ultimately reaching a relatively low level. This outcome suggests that the chosen training strategy, hyperparameter configuration, and optimization techniques are effective, allowing the model to learn the data distribution efficiently while maintaining strong generalization performance. Additionally, the stabilization of the loss curves indicates that the training process has effectively converged, suggesting that training can be halted or fine-tuned further to ensure optimal performance on the test set. Finally, this paper also gives the T-SNE results after training, as shown in Figure 3.

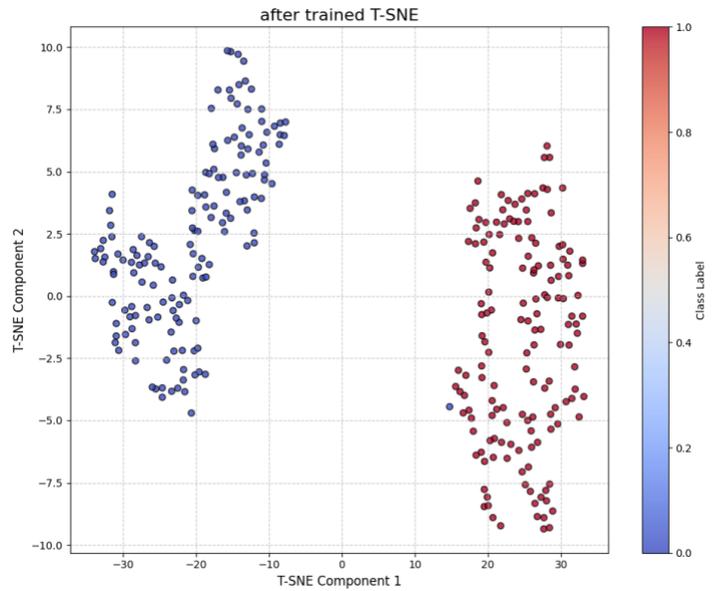

Figure 3. T-SNE result map after training

From the T-SNE results, it is evident that after training, the data points form distinct cluster-like distributions in the two-dimensional space, indicating that the model has successfully learned the feature differences between different classes. As observed in the visualization, the two categories (represented in blue and red) are well separated, suggesting that the model has developed strong discriminative capabilities in the high-dimensional feature space. The presence of a clear boundary between the classes demonstrates that the model effectively extracts distinguishing features without causing sample overlap, thereby validating its effectiveness.

Furthermore, the overall data distribution demonstrates that the T-SNE dimensionality reduction retains intra-class compactness while ensuring inter-class separability. The blue and red data points are well-clustered in distinct regions without significant overlap, indicating that the model effectively distinguishes between different categories in the feature space. Even when dealing with an imbalanced dataset, the model successfully learns the distribution patterns of the minority class.

However, while the T-SNE results illustrate a clear class separation, further quantitative evaluation is necessary to assess the robustness of the classification boundaries. For instance, if significant distribution shifts occur in certain test data samples, it may indicate that the model is still susceptible to overfitting. Additionally, since T-SNE is a nonlinear dimensionality reduction method, it may exaggerate the separation between classes, meaning that the actual decision boundaries in the high-dimensional space may not be as well-defined as they appear in the visualization. Therefore, a comprehensive evaluation incorporating classification metrics such as Precision, Recall, and AUC is essential to fully validate the model's generalization performance.

## IV. Conclusion

This study proposes an imbalanced data classification method based on deep probabilistic graphical models (DPGMs) and validates its effectiveness through experiments on a credit card fraud detection dataset. The experimental results demonstrate that the proposed method outperforms traditional oversampling techniques, ensemble learning approaches, and attention-based models in key metrics such as AUC and F1-score, confirming the effectiveness of probabilistic modeling in handling imbalanced classification tasks. By integrating variational inference, class-weight adjustment, and adversarial learning mechanisms, our model more accurately captures the feature distribution of the minority class, enhancing the classifier's discriminative ability while mitigating the overfitting issues commonly observed in traditional methods.

Despite the promising performance of our approach in imbalanced data classification, several aspects warrant further improvement. For instance, in cases of extreme imbalance, the minority class samples may still provide insufficient information, potentially limiting the model's generalization capability. Additionally, deep probabilistic graphical models involve high computational complexity, requiring extensive sampling and variational inference steps during training, which may impact deployment efficiency. Therefore, future research could focus on optimizing the computational efficiency of probabilistic modeling to enhance the model's adaptability across different data distributions. Several directions can be explored in future research. More efficient Bayesian optimization methods can be investigated to reduce the computational cost of DPGMs, making them applicable to larger-scale imbalanced datasets. Furthermore, in practical applications, federated learning frameworks can be incorporated to enable cross-institutional model training while preserving data privacy, thereby enhancing the applicability of imbalanced classification methods in real-world scenarios.